\documentclass[runningheads]{ECML-formatting/llncs}
\usepackage{graphicx}
\usepackage{amsmath}
\usepackage{amsfonts}
\usepackage{amssymb}
\usepackage{xcolor}
\usepackage{booktabs}
\usepackage{multirow}
\usepackage{tabularx}

\newcommand{\PA}[1]{\textbf{\textcolor{red}{[#1]}}}

\begin{document}
\title{Cooperative Bayesian Optimization \\for Imperfect Agents
}
%
%

\author{Ali Khoshvishkaie\inst{1} \and Petrus Mikkola\inst{1} \and Pierre-Alexandre Murena\inst{1,2} \and Samuel Kaski\inst{1,3}}

\authorrunning{A. Khoshvishkaie et al.}
%

\institute{Department of Computer Science, Aalto University, Helsinki, Finland \and Hamburg University of Technology, Hamburg, Germany \and Department of Computer Science, University of Manchester, Manchester, UK \\
\email{\{firstname.lastname\}@aalto.fi}}

\toctitle{Cooperative Bayesian Optimization for Imperfect Agents}
\tocauthor{Ali~Khovishkaie, Petrus~Mikkola, Pierre-Alexandre~Murena, Samuel~Kaski}

\maketitle              

\begin{abstract}
  We introduce a cooperative Bayesian optimization problem for optimizing black-box functions of two variables where two agents choose together at which points to query the function but have only control over one variable each. This setting is inspired by human-AI teamwork, where an AI-assistant helps its human user solve a problem, in this simplest case, collaborative optimization. 
We formulate the solution as sequential decision-making, where the agent we control models the user as a computationally rational agent with prior knowledge about the function. We show that strategic planning of the queries enables better identification of the global maximum of the function as long as the user avoids excessive exploration. This planning is made possible by using Bayes Adaptive Monte Carlo planning and by endowing the agent with a user model that accounts for conservative belief updates and exploratory sampling of the points to query.

\end{abstract}

\section{Introduction}
Human-AI cooperation refers to the collaboration between human and artificial intelligence (AI) driven agents to achieve a common goal~\cite{o2022human}. In the \emph{cooperative} scenario, the agents work autonomously but interdependently, each leveraging their unique skills and abilities to collectively reach the shared objective. 
The cooperation between a human and an AI agent can be impaired by limitations in their information processing abilities and various other factors such as biases, heuristics, and incomplete knowledge~\cite{helander2014handbook}. It has already been established that any cooperation is more effective when the involved agents have a \emph{theory of mind} of the others \cite{etel2019theory}. 
It would therefore be helpful if the AI agent could take into account the human's information processing capabilities and biases and adapt to the changing needs and preferences of the human user~\cite{sears2009human}.

\begin{figure}[t]
    \centering
    \includegraphics[width=\textwidth]{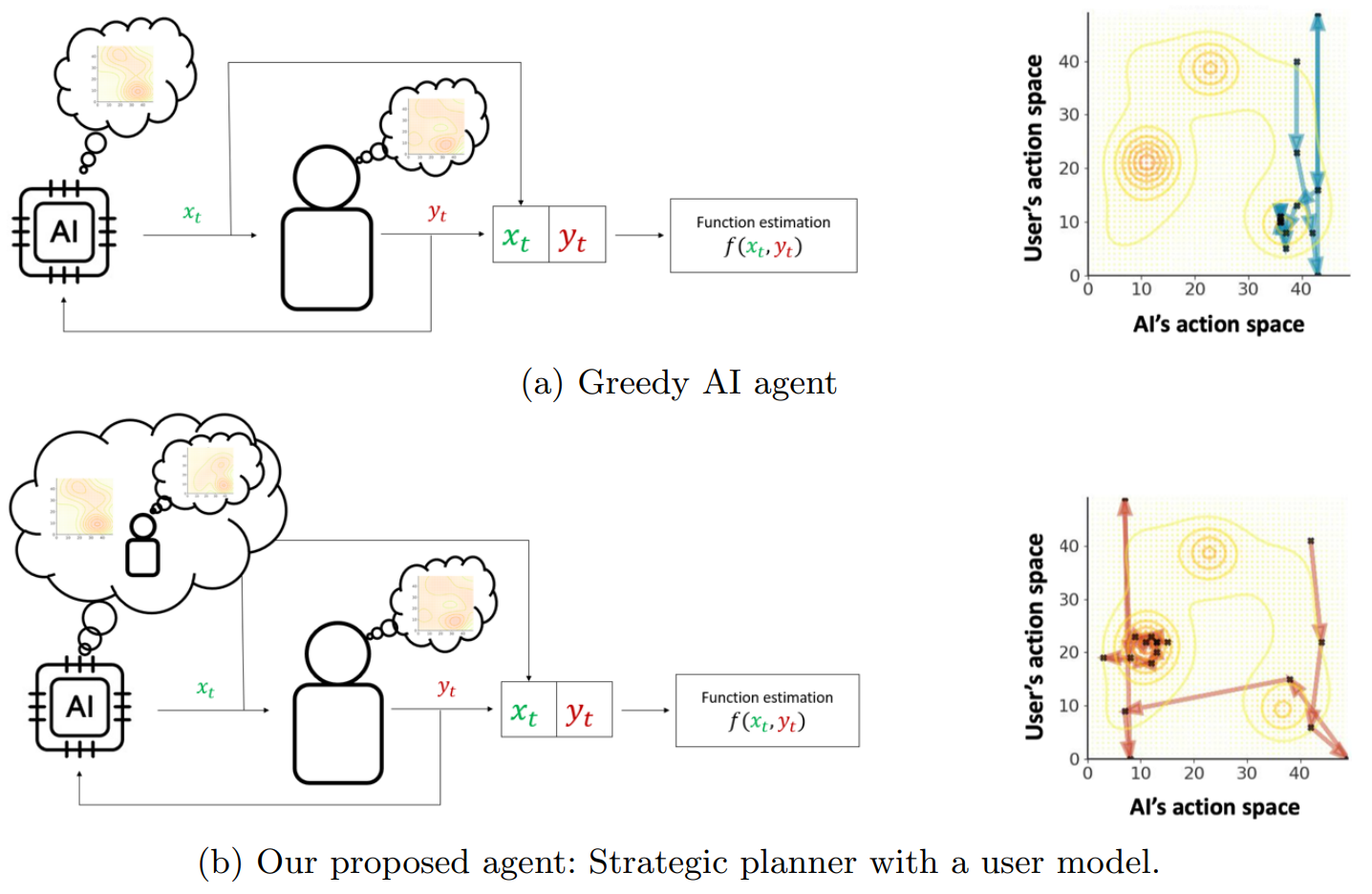}
    \caption{Interaction scenario between the user and the AI agent in the optimization task. Unlike a \emph{greedy agent} (a), the AI agent we propose (b) has a \emph{model} of the user and plans its actions by anticipating the user's behaviour. This results in a more efficient cooperative exploration of the domain, and therefore avoids getting stuck in a local optimum. This is visible in the right-hand side plots, showing the corresponding trajectories of queries to the function $f$.}
    \label{fig:protocol}
\end{figure}

A specific kind of human-AI cooperation is when the decision is jointly taken by two agents for a common goal, and each controls only their part of the decision.
An illustrative example is Hand and Brain chess, a team chess variant in which two players (the Hand and the Brain) play on each side. Each move is jointly decided by the team, with the Brain calling out a piece and the Hand being responsible for moving it. In this game, the Brain should essentially consider a move that is understandable for the Hand. Otherwise, the Hand moves the piece to a strategically bad position, resulting in a disastrous move. If each player carries out their task without anticipating the other team member, the team will end up taking a sub-optimal action. The anticipation is done by building a \emph{model} of the partner.

To study this setup in a controlled environment, we propose a cooperative Bayesian optimization task. The AI agent and human user aim to perform a sequential black-box optimization task in a 2D space. 
At each step, the human-AI team chooses a point to query the function. The choice is made by the AI agent opting for the first coordinate and then the human user selecting the other one. In this optimization task, the human user and the AI agent, both with partial information, cooperatively take part in data acquisition. We formulate this cooperative data acquisition as a repeated Bayesian game between the user and the agent played for a finite horizon.
The contributions of this paper are:
\begin{itemize}
    \item We propose a collaborative AI algorithm for settings where the AI agent plans its action by assessing the user's knowledge and decision process without any prior interaction with the user.
    \item We show empirically that the algorithm is able to learn the user's behaviour in an online setting and use it to anticipate the user's actions. 
    \item We show empirically that the algorithm helps the team in the optimization task (measured as the team optimization score) compared to various baselines, such as a greedy algorithm that maximizes its own beliefs. This is done by helping a better exploration of the domain of the function.
\end{itemize}

\section{Cooperative Bayesian Optimization}
\subsection{Problem Formulation}

We consider a problem where a team of two agents, the \emph{human user} and the \emph{AI agent}, aims to maximize a black-box function $f: \mathcal{X} \times \mathcal{Y}$ of two parameters $(x, y) \in \mathcal{X} \times \mathcal{Y}$. Note here that the function is not necessarily 2-dimensional. The team explores the domain $\mathcal{X} \times \mathcal{Y}$ by acquiring new observations of~$f$. 
The exploration consists of a sequence of queries of~$f$ at points $(x_t, y_t) \in \mathcal{X} \times \mathcal{Y}$. The outcomes of the query are noisy and we denote by~$\bar{f}(x,y)$ the outcome of the query at point $(x,y)$. In this respect, the task of the team is similar to a \emph{Bayesian Optimization} (BO) task. 

The team proceeds by sequentially querying $T$~points. At each step~$t$, the team adopts the following protocol for the choice of $(x_t, y_t)$, presented in Figure~\ref{fig:protocol}. The AI agent selects~$x_t \in \mathcal{X}$ first. The human user observes the value of~$x_t$ picked by the AI agent and then selects~$y_t \in \mathcal{Y}$. Finally, both agents observe the selected tuple $(x_t, y_t)$ and the value of $f(x_t, y_t)$. In this paper, we adopt the point of view of the AI agent and therefore focus on how to optimally select the first coordinate~$x_t$. It is important to mention that $\mathcal{X}$ and $\mathcal{Y}$ are not necessarily one-dimensional, but can describe any two sets of variables.


The final performance of the optimization process is measured by the \emph{optimization score} (described in Section~\ref{subsec:experiments}). We view this score as a more understandable alternative to the directly related measure of simple regret, defined as $f^* - f^*_T$.


\subsection{Mathematical Formalization}
\label{subsec:formalization}

We address the problem of the AI agent as a repeated Bayesian game, using the formalism of model-based reinforcement learning, considering the AI agent as a decision-making agent interacting with an environment made up of the function~$f$ and the human user. In this environment, the agent takes actions (choice of a coordinate~$x_t$) and gets rewarded depending on the action $x_t$, the user's choice $y_t$ and the value of the function $f(x_t, y_t)$.

We describe the agent's decision-making problem as Partially Observable Markov Decision Process (POMDP) $\mathcal{M} = \langle \mathcal{A}, \mathcal{S}, \mathcal{T}, \Omega, \mathcal{O}, \mathcal{R} \rangle$, where the notations are explained in what follows.

The space~$\mathcal{A}$ is the space of the actions available to the agent. In our context, it corresponds to the set $\mathcal{X}$ of points available to the agent. For this reason, we will use~$x$ (instead of the standard notation~$a$ usually used in POMDPs) to designate the coordinate chosen by the agent. A state $s \in \mathcal{S}$ describes a state of the agent's environment, which is made up of the function $f$ and the user. A state is then defined as a tuple $s = (f_{AI}, \theta)$, where $f_{AI}: \mathcal{X} \times \mathcal{Y} \rightarrow \mathbb{R}$ is agent's estimation of function $f$ and $\theta$ is a parameters set characterizing the user. The transition $\mathcal{T}(s, x, s^\prime)$ measures the probability of a transition from state $s$ to state $s^\prime$ after agent's action $x$. 
By definition, the function $f$ is fixed, and consequently the transition probability~$\mathcal{T}$ can be written as:
\begin{equation}
    \mathcal{T}(s, x, s^\prime) = \mathbb{I}(f_{AI} = f_{AI}^\prime) p(\theta^\prime | s, x)
    \label{eqn:transition-proba}
\end{equation}
where $s = (f_{AI}, \theta)$, $s^\prime = (f_{AI}^\prime, \theta^\prime)$ and $\mathbb{I}$ is the identity function.
An observation $\omega \in \Omega$ corresponds to the user's choice~$y$ and the value of $f$ at point $(x,y)$, i.e. $\Omega = \mathcal{Y} \times \mathbb{R}$. The observation prediction $\mathcal{O}(\omega | s, x) \in [0, 1]$  is the probability that $\omega \in \Omega$ is observed after action $x$ has been played within environment state~$s$. Unlike some other settings, the observation prediction~$\mathcal{O}$ in our context does not depend on the new state, but only on the state before the action. This probability decomposes as
\begin{equation}
    \mathcal{O}(\omega=(y, z) | s, x) = p(y | s, x) p(\bar{f}(x,y) = z | s, x, y)
    \label{eqn:observation-proba}
\end{equation}
where the probability $p(y | s, x)$ corresponds to user's decision-making and the probability $p(\bar{f}(x,y) = z | s, x, y)$ to function sampling. The reward~$\mathcal{R}(x, s, \omega)$ measures the pay-off of agent's action~$x$ in state~$s$ after observing~$\omega$. The choice of the reward function in our implementation will be discussed in Section~\ref{subsec:reward}.


\subsection{User Model}
\label{subsec:user-model}

In Equations~\ref{eqn:transition-proba} and \ref{eqn:observation-proba}, the probabilities $p(\theta^\prime | s, x)$ and $p(y | s, x)$ describe the user's behaviour, that is, how the user updates their beliefs and how they make decisions. We note that $p(\theta^\prime | s, x)$ can be decomposed as
\begin{equation}
    p(\theta^\prime | s, x) = \int_z \sum_y \  p(\theta^\prime | s, x, y, f(x, y) = z)
    p(f(x, y) = z | s, x, y) p(y | s, x) dz
    \label{eqn:usermodel-proba}
\end{equation}
where the term $p(f(x, y) = z | s, x, y)$ does not depend on the user. Therefore, the user's behaviour is fully defined by $p(\theta^\prime | s, x, y, f(x, y))$ and $p(y | s, x)$. 

In the following, we will call the tuple $( p(\theta^\prime | s, x, y, f(x, y)), p(y | s, x) )$ the \emph{user model}. The user model describes the role played by the user within the environment of the agent. In practice, it will be used to simulate the behaviour of the user, which is useful in particular when planning for the action to play. The user model is not necessarily an accurate description of the user's behaviour, but is a model used by the agent for making decisions. The choice of this model will restrict the possibilities of behaviours that the agent will be able to consider. In the case where the user is human, a useful user model should be able to describe computationally rational behaviours~\cite{gershman2015computational}.

\section{Implementation}
In this section, we introduce the practical solution to the Cooperative Bayesian Optimization problem, considering a minimal user model. This model describes a user with partial knowledge about the function, able to update their belief and select their actions in a way that balances exploitation and exploration.

\subsection{Bayes Adaptive Monte Carlo Planning}
\label{subsec:planning}

In order to solve the POMDP introduced in Section~\ref{subsec:formalization} and plan the AI agent's actions, we rely on a Bayesian model-based Reinforcement learning method. This method is used to perform a \emph{zero-shot} planning, where the agent has no initial information about the user's behaviour.
At each iteration, the model is updated based on the previous user's actions, and a zero-shot planning method is employed to plan for the future.

In order to solve the POMDP, the posterior distribution of the parameters is estimated using the inference method described in Section~\ref{subsec:inference} below. This posterior distribution is used to plan the actions~$x_t$ by enabling a Monte-Carlo estimation of the value of each action: At each iteration, we run several simulations with fixed state $s_t$ sampled from the posterior distribution. In these conditions, having a fixed and known state transforms the POMDP into a simple MDP: This makes it possible to compute the value of the action for this state and, consequently, to get a Monte-Carlo estimation of the value of an action. Finally, the action that maximizes the estimated value is chosen. It has been proven that this process converges to the Bayes-optimal policy with infinite samples~\cite{guez2013scalable}.

\subsection{User Model Specification}
\label{subsec:user-model-spec}

We propose a simple user model describing a computationally rational user with partial knowledge about the function to be optimized. This user model is an instantiation of the general form of user models as introduced in Section~\ref{subsec:user-model}.

\paragraph{User's Knowledge.}
We represent the user's partial knowledge of the function~$f$ using a Gaussian Process~\cite{rasmussen2006gp}. A Gaussian Process (GP) is a stochastic process over real-valued functions, such that every finite collection of these random variables has a multivariate normal distribution. We will denote this GP at step~$t$ as $f_{um}^{(t)}$. We emphasize that $f_{um}$ is not a function, but a prior over functions~$\mathcal{X} \times \mathcal{Y} \rightarrow \mathbb{R}$. This choice is motivated by the observation that Bayesian Optimization based on GPs provides a surprisingly good framework to explain active function learning and optimization in humans~\cite{borji2013}.

The user's GP is assumed to have been initialized based on the observation of a collection of~$N_u$ points~$\mathcal{D}_u = \lbrace (x_i^u, y_i^u, \bar{f}(x_i^u, y_i^u)) \rbrace_{i = 1, \dotsc, N_u}$, using GP regression. For any unseen function value $f(x,y)$, GP regression models this as a Gaussian random variable with closed-form mean and variance (see \cite{rasmussen2006gp}, Equations (2.23) and (2.24)). The equations require specifying the covariance (kernel) function, which in this paper is taken to be the squared exponential kernel \cite[Eq. (2.16)]{rasmussen2006gp}. The hyperparameters of the kernel function are optimized by maximizing the marginal likelihood.

\paragraph{Belief Update.}

The values of the function~$f$ sampled during the interaction are observed by the user and used to sequentially update their GP~$f_{um}$. At time~$t$, the user's GP~$f_{um}^{(t)}$ is updated by observing $\mathcal{D}_t = \{(x_t,y_t), \bar{f}(x_t,y_t)\}$. 
We denote by $\mathcal{B}_{bayes}(f_{um}^{(t)}|\{(x_t,y_t),\bar{f}(x_t,y_t)\})$ the GP obtained after Bayes optimal belief updating, defined as the standard updates (Equations (2.23) and (2.24) in \cite{rasmussen2006gp}). 

However, it has been documented in behavioural studies \cite{el1995people} that humans deviate from the Bayesian optimal belief update, because of various cognitive biases~\cite{tversky1974judgment}. Consequently, in our user model, we consider the \emph{conservative belief updating} operator $\mathcal{B}$ introduced by Kovach~\cite{kovach2021conservative}:
\begin{equation}
    f_{um}^{(t+1)} = \alpha f_{um}^{(t)} + (1-\alpha)\mathcal{B}_{bayes}(f_{um}^{(t)}|\{(x_t,y_t),\bar{f}(x_t,y_t)\}),
    \label{eqn:conservative-belief-update}
\end{equation}
where $\alpha \in [0,1]$ represents the degree of \emph{conservatism}. A low values of~$\alpha$ corresponds to an almost Bayes-optimal behaviour, while the case~$\alpha = 1$ corresponds to the user ignoring the new observations and not updating their belief.

\paragraph{Decision-Making.}

Motivated by the observation of Borji and Itti~\cite{borji2013}, we model the user's choice of an action $y_t$ as the maximization of an acquisition function $y \mapsto A(x_t,y)$. We consider the UCB acquisition function based on the GP $f_{um}^{(t)}$:
\begin{equation}
    A_{t}(y|x_{t}) = \mathbb{E}\left[f_{um}^{(t)}(x_{t}, y) \right] + \beta \sqrt{\mathbb{V}\left[f_{um}^{(t)}(x_{t},y) \right]}
\end{equation}
where $\mathbb{E}[f_{um}^{(t)}(x_{t}, y) ]$ and $\mathbb{V}[f_{um}^{(t)}(x_{t}, y)]$ are respectively the mean and the variance of the GP~$f_{um}^{(t)}$ at point $(x_t, y)$, 
and $\beta \in [0,1]$ is an \emph{exploration-exploitation trade-off} parameter. A low value of $\beta$ corresponds to less explorative behaviour, exploiting the current belief over~$f$, while a larger value corresponds to more explorative behaviour, evaluating~$f$ at points with larger uncertainty. Given the AI's action $x_{t}$, a sensible choice of an action $y_{t}$ for the user would consist in maximizing the acquisition function $A_{t}(y|x_{t})$.

This choice of~$y_t$ by a maximization can be interpreted as an event of many pairwise comparisons among different actions $y \in \mathcal{X}$: Choosing the action $y_{t}$ means preferring it to all the others $y \neq y_{t}$. Inspired by \cite{PPBO}, we build a probabilistic model of user preferences upon Thurstone’s law of comparative judgment~\cite{thurstone1994law} by assuming that the user's action $y_{t}$ given $x_{t}$ is corrupted by Gaussian noise,
\begin{equation}
    y_{t} = \text{arg}\max_{y}\left( A_{t}(y|x_{t}) + W(y) \right),
    \label{eqn:a_max}
\end{equation}
where $W$ is a white Gaussian noise with mean $\mathbb{E}[W(y)]= 0$ and auto-correlation 
$\mathbb{E}[W(y)W(y')] =\sigma^2$ if $y = y'$ 
and 0 otherwise. The likelihood~$p(y_{t}|s_t, x_{t})$ of a single observation $y_{t}|x_{t}$ corresponding to this noise process takes the form
\begin{equation}
p(y_{t}|s_t, x_{t}) = \prod_{i=1}^{m} \left(1 - [\Phi * \phi] \left(\frac{A_{t}(y_i|x_{t}) - A_{t}(y_{t}|x_{t})}{\sigma}\right)\right),
\label{eqn:decision-likelihood}
\end{equation}
where $\Phi$ and $\phi$ are the cumulative and density function of the standard
normal distribution, respectively, $*$ and is the convolution operator.
To evaluate the likelihood, $f_{um}^{(t)}$ and $A_{t}(y|x_{t})$ should be computed recursively by using the aforementioned equations. For fixed $\alpha$ and $\beta$, this is possible given the function sampling data $(\mathcal{D}_t)_{t=1}^T$. The joint likelihood $P\left((y_t | x_t)_{t=1}^T,(\mathcal{D}_t)_{t=1}^T \big| \alpha,\beta \right)$ is the product of the single events $y_t | x_t$ for $t = 1,...,T$. 

\paragraph{Summary: Definition of the User Model.} The introduced user model is characterized by three parameters: the user's knowledge of the function $f_{um}$, the degree of conservatism~$\alpha$ and the degree of explorativeness~$\beta$. Using the notations of Section~\ref{subsec:formalization}, we can write~$\theta = (f_{um}, \alpha, \beta)$. We notice that these parameters are of different natures though: $\alpha$ and $\beta$ are characteristics of the user, while $f_{um}$ corresponds to a \emph{mental state}, i.e. a description of what the user knows. 

When defining the belief-updating probability $p(\theta^\prime | s, a, y, f(a, y))$, we assume that the parameters~$\alpha$ and $\beta$, as characteristics of the user, are stationary and therefore are not updated during the interaction. Only the user's GP is updated, following Equation~\ref{eqn:conservative-belief-update}. With our definition of this user model, the user's decision-making~$p(y | s, a)$ is defined in Equation~\ref{eqn:decision-likelihood}.

\subsection{Inference of the User Model Parameters}
\label{subsec:inference}

The parameters~$\theta = (f_{um}, \alpha, \beta)$ are not observed and need to be estimated online during the interaction, based on the user's actions. 
We adopt a Bayesian approach and the inference consists of estimating, at each time 
step~$t$, the posterior distribution $p\left(\alpha, \beta, f_{um} \big|\ (y_\tau | x_\tau)_{\tau=1}^t, (\mathcal{D}_\tau)_{\tau=1}^t  \right)$  given the interaction data $(y_\tau | x_\tau)_{\tau=1}^t$ and the function sampling data $(\mathcal{D}_\tau)_{\tau=1}^t$ with $\mathcal{D}_\tau = (x_\tau, y_\tau, \bar{f}(x_\tau, y_\tau))$. For this, we use the following decomposition:
\begin{align*}
    p & \left(\alpha, \beta, f_{um} \big|\ (y_\tau | x_\tau)_{\tau=1}^t, (\mathcal{D}_\tau)_{\tau=1}^t  \right) = \\ 
    & \qquad p\left(\alpha, \beta \big|\ (y_\tau | x_\tau)_{\tau=1}^t, (\mathcal{D}_\tau)_{\tau=1}^t  \right) p\left(f_{um} \big|\ (y_\tau | x_\tau)_{\tau=1}^t, (\mathcal{D}_\tau)_{\tau=1}^t, \alpha, \beta \right)
\end{align*}

\paragraph{Estimation of $(\alpha, \beta)$.} The estimation of $(\alpha,\beta)$ is done using Bayesian belief update. The initial prior is chosen to be the uniform distribution over the unit cube. 
The posterior distribution is approximated using the Laplace approximation, which consists in the following. The maximum a posteriori (MAP) estimate $(\alpha_{\textrm{MAP}},\beta_{\textrm{MAP}})$ is computed by numerically maximizing the log posterior with the BFGS algorithm, which also approximates the Hessian. The posterior $p\left(\alpha,\beta \big|\ (y_\tau | x_\tau)_{\tau=1}^t, (\mathcal{D}_\tau)_{\tau=1}^t  \right)$ is approximated as a Gaussian distribution centered on $(\alpha_{\textrm{MAP}},\beta_{\textrm{MAP}})$ with the covariance matrix corresponding to the inverse of the negative Hessian at the MAP estimate.

\paragraph{Estimation of $f_{um}$.} The update of $f_{um}$ as given in Equation~\eqref{eqn:conservative-belief-update} is deterministic when $\alpha$ is given. Consequently, the term $p\left(f_{um} \big|\ (y_\tau | x_\tau)_{\tau=1}^t, (\mathcal{D}_\tau)_{\tau=1}^t, \alpha, \beta \right)$ is trivial and does not need to be computed during the interaction. Since our planning algorithm (described in Section~\ref{subsec:planning}) relies on sampling from the parameters~$(\alpha, \beta, f_{um})$, $f_{um}$ is computed from the whole trajectory~$(\mathcal{D}_\tau)_{\tau=1}^t$ using the sampled value of~$\alpha$. For the initialization $f_{um}^{(0)}$, we consider that the user has a uniform prior over the function. This interprets as ignoring the fact that the user has prior knowledge. 

\subsection{Choice of the Reward Function}
\label{subsec:reward}

The reward for the agent, as introduced in Section~\ref{subsec:formalization}, is designed to be a compromise of two parts, exposed in the following. 

The first part is the expectation of the UCB score over the user's future action, calculated with $f_{um}$ as estimated in the user model:  
\begin{equation}
    \mathcal{R}_{1}(x, s, \omega) = {\mathbb{E}}_{y \sim A_{usr}}\left[ UCB(x, y) \right]
    \label{eqn:R1}
\end{equation}
Intuitively, this first part~$\mathcal{R}_1$ shows how desirable the point $(x, y)$ is for the user when the AI selects $x$. Therefore it values actions $x$ for which the user is able to find a reasonably good $y$ to query the function. Since~$\mathcal{R}_1$ is based on the UCB score, it also guarantees a trade-off between the exploration and exploitation of the query point.

When the user's behaviour is almost uniform over actions (e.g. when the user is more explorative, because of having little knowledge of $f$ or because of a high~$\beta$), reward~$\mathcal{R}_1$ is close to constant and is not enough to make good choices of~$x_t$. We solve this problem by introducing a second part in the reward definition, that is based on the AI agent's knowledge of the function. This reward is defined as the average UCB score over the top K promising $y$ values upon the AI's knowledge for a chosen action $x$:
\begin{equation}
    \mathcal{R}_{2}(x, s, \omega) = \frac{1}{K}\sum_{y \in top_K(A_{ai})} UCB(x, y) 
\end{equation}
This reduces the risk of relying too much on the user model, which is not prefect, especially at the beginning of the interaction. 

We define the total reward as a linear combination of these two components: 
\begin{equation}
    \mathcal{R}(x, s, \omega) = \mathcal{R}_{1}(x, s, \omega) + C \  \mathcal{R}_{2}(x, s, \omega)
    \label{eqn:total-reward}
\end{equation}
where $C$ is a compromising factor between the two terms, a hyperparameter of the proposed method.

\section{Empirical Validation}
 In this section, we study the performance of our method in the proposed cooperative Bayesian game (Section 2). We examine scenarios where prior information is unevenly distributed among agents and when the human user characteristics vary. In particular, we are interested in how the user's degree of conservatism and explorativeness affect the outcome of the Bayesian game.\footnote{Implementation of our method and source code for the experiments are available at \url{https://github.com/ChessGeek95/AI-assisted-Bayesian-optimization/}.}

\subsection{Experimental Setup}

\paragraph{Domain.} We choose as a function~$f$ a 3-modal variant of the \emph{Himmelblau function}. It is defined on $[0,1]^2$ and has 3 minima, located respectively at $(0.46, 0.8)$, $(0.22, 0.44)$ and $(0.74, 0.18)$. The amplitude of the maxima can be adjusted.

\paragraph{Experimental protocol.} 
We consider a synthetic user whose characteristics can be controlled. The agent follows the specification of a computationally rational user presented in Section~\ref{subsec:user-model-spec}: we assume a user who follows a Bayesian optimization routine based on the UCB acquisition function with an explorativeness parameter $\beta$, and a conservative GP-based belief updating with a conservatism parameter $\alpha$. We create $2\times2$ configurations of the human user characteristics by considering the possible combinations of the values $\alpha \in \{0.1,0.6\}$ and $\beta \in \{0.2,0.7\}$. For example, the configuration $\alpha = 0.1$ and $\beta = 0.7$ refers to a human user who is conservative in belief updating but explorative in decision-making. These values have been chosen to reflect the extremes, with the user being almost completely conservative or almost perfectly Bayesian, and the user being almost exclusively exploitative or almost exclusively explorative.

We study the impact of this prior information by considering $3\times3$ configurations of prior information (see Section~\ref{subsec:formalization}) as follows. We provide each of the two agents with either $N=5$ points around local maxima or the global maximum, or no prior functions evaluations at all. 
We use the terms ``\texttt{Local}", ``\texttt{Global}", and ``\texttt{None}" to refer to these configurations by considering possible permutations: (AI's prior, human's prior).
The points are drawn from a multi-normal distribution centered on the position of the maximum (local or global).
For example, (\texttt{Global}, \texttt{Local}) refers to the configuration, where the AI agent has $N=5$ prior points around the global maximum, while the human user agent has $N=5$ points around local maxima of the function.

For the experiments, we consider a discretization of the domain~$\mathcal{X} \times \mathcal{Y}$ into a $50 \times 50$ grid. Given a simulated user, each experiment consists of 20 interaction steps. 
The results are averaged over a sample of 3 different functions~$f$, generated as described above, and 10 different prior samples (initial points available to each agent before the interaction, see Section~\ref{subsec:user-model}). For our agent, we use the reward defined in Equation~\ref{eqn:total-reward} with $C = 1$. 

All experiments were run on a private cluster consisting of a mixture of Intel$^\text{\textregistered}$ Xeon$^\text{\textregistered}$ Gold 6248, Xeon$^\text{\textregistered}$ Gold
6148, Xeon$^\text{\textregistered}$ E5-2690 v3 and Xeon$^\text{\textregistered}$ E5-2680 v3 processors.

\paragraph{Baselines.} To investigate the strengths of our method, we compare it to four baselines, two of which correspond to a single-agent Bayesian Optimization. 

The single-agent BO baselines correspond to one single agent making the decision, i.e., opting for both coordinates of the point to query, and therefore correspond to the standard BO problem. The baselines illustrate empirical lower and upper bounds of the optimization performance:
\begin{itemize}
    \item[-] \texttt{VanillaBO (random)}: Single-agent baseline, querying points~$(x, y)$ uniformly at random on the domain~$\mathcal{X} \times \mathcal{Y}$. This is equivalent to two agents querying coordinates randomly, which is a lower bound on the performance that any team should at least achieve.
    \item[-] \texttt{VanillaBO (GP-UCB)}: Single-agent baseline, querying points~$(x, y)$ using an upper confidence bound~\cite{cox1992statistical} score upon a Gaussian processes pre-trained on the prior points. Since the agent has access to all prior data and absolute control over both coordinates, this is an upper bound on the performance. The value of $\beta$ for this agent is chosen to be $\beta = 0.05$: it has been chosen because it gives optimal results compared to other $\beta$.
\end{itemize}

We also compare the performance of our method to two other comparable multi-agent BO algorithms, corresponding to different strategies for solving the Cooperative Bayesian Optimization task:
\begin{itemize}
    \item[-] \texttt{RandomAI}: The AI agent chooses~$x$ uniformly at random on the domain~$\mathcal{X}$.
    \item[-] \texttt{GreedyAI}: The AI chooses~$x$ by picking the first coordinate of the UCB score maximizer. It maximizes its own utility function (UCB score) without considering the other agent, hence the name. As for the \texttt{GP-UCB} agent, the value of $\beta$ for this agent is also chosen to be $\beta = 0.05$, 
\end{itemize}

\subsection{Experiments}
\label{subsec:experiments}

\subsubsection{Experiment 1: Evolution of the optimization performance.}
We first study the efficiency of our algorithm in helping the team in the optimization task. To do so, we introduce, as a metric, the \emph{optimization score}. We define this score as the maximum function value $f_t^*$ queried during the cooperative game of $t$ rounds. Since the objective function is normalized between 0 and 100, an optimization score of 100 denotes maximum performance (also note that $\textrm{simple regret}=100-\textrm{optimization score}$). 
\begin{figure}
    \centering
    \includegraphics[width=.6\textwidth]{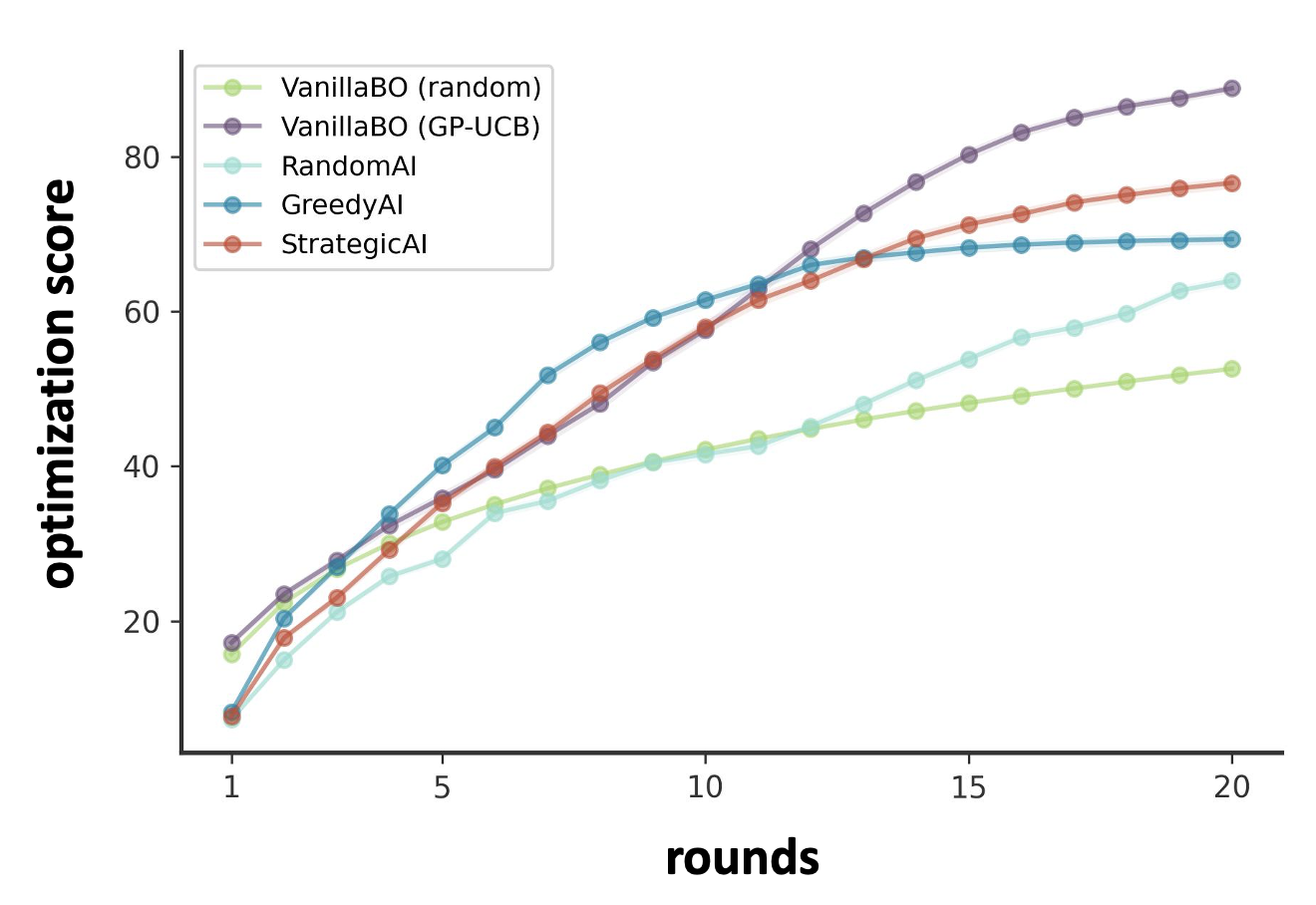}
    \caption{Evolution of the optimization performance during the interaction. At the end of the interaction, our agent (\texttt{StragicAI}) gets better performance than other baselines. It performs slightly worse than the \texttt{VanillaBO (GP-UCB)}, because, unlike this baseline, the \texttt{StrategicAI} does not have control over the full domain~$\mathcal{X} \times \mathcal{Y}$.}
    \label{fig:mainresult}
\end{figure}

The evolution of the optimization performance over the optimization rounds is presented in Figure~\ref{fig:mainresult}. It can be seen that our method indeed reaches better performance compared to the \texttt{GreedyAI} and \texttt{RandomAI} baselines. However, in the initial rounds, \texttt{GreedyAI} displays much better performances (even better than the \texttt{VanillaBO (GP-UCB)} agent): this is because \texttt{GreedyAI} exploits prior information and therefore is quickly able to guide the user toward finding a local maximum. However, once the optimum is found, it does not explore further and does not find any global optimum, unlike our method, which is more explorative from the beginning. We also notice that the \texttt{RandomAI} has initial performance close to the random VanillaBO baseline, but keeps improving: this is due to the fact that this agent keeps exploring, but in a sub-optimal way. Finally, we still notice that the \texttt{VanillaBO (GP-UCB)} baseline is indeed a valuable upper-bound in the long-term: even though our \texttt{StrategicAI} has similar performances on the first rounds, the AI not having total control over the exploration ends up making slightly less optimal decisions.

\subsubsection{Experiment 2: Impact of the user's parameters.}\label{exp_user_parameters}

The results presented in Figure~\ref{fig:mainresult} are averaged over all user parameters. 
To study the impact of the user's conservatism and explorativeness on the optimization performance, we exploit the possibility offered by a controlled synthetic user to directly interpret the performance of our method in the case of various user profiles.
The final optimization score for different $(\alpha,\beta)$ configurations is reported in Table~\ref{tab:impact-user-parameters}. The scores are averaged over all combined prior knowledge configurations. These results confirm that the AI's strategic planning significantly improves optimization performance in all scenarios when compared to greedy or random strategies, but with the highest margin for conservative users. This suggests that strategic planning is more crucial when users update their beliefs conservatively. In contrast, the level of user exploration does not significantly affect the size of the margin.  

As an addition to this experiment, we performed an ablation study to check the role played by the choice of the reward (Equation~\ref{eqn:total-reward}), comparing the cases where $C=1$ (used in all other reported experiments) and where $C = 0$ (which corresponds to reward~$\mathcal{R}_1$ introduced in Equation~\ref{eqn:R1}). The results reveal that the performance of strategic AI deteriorates with an explorative user, by using $\mathcal{R}_1$ instead of the full reward~$\mathcal{R}$ (which corresponds to the case $C = 0$).


\begin{table}[]
\caption{Impact of the user's conservativeness ($\alpha$) and explorativeness ($\beta$) onto the optimization score.}
\label{tab:impact-user-parameters}
\setlength{\tabcolsep}{3pt}
\def\arraystretch{1.2}
\centering
\begin{tabular}{c|cccc|}
\cline{2-5}
 & \multicolumn{2}{c|}{$\beta = 0.2$} & \multicolumn{2}{c|}{$\beta = 0.7$} \\ \cline{2-5} 
 & \multicolumn{1}{c|}{$\alpha = 0.1$} & \multicolumn{1}{c|}{$\alpha = 0.6$} & \multicolumn{1}{c|}{$\alpha = 0.1$} & $\alpha = 0.6$ \\ \hline
\multicolumn{1}{|c|}{GP-UCB} & \multicolumn{4}{c|}{$ 88.9 \pm 21.4 $} \\ \hline
\multicolumn{1}{|c|}{StrategicAI, $C=1$ (ours)} & \multicolumn{1}{c|}{$77.5 \pm 24.8$} & \multicolumn{1}{c|}{$76.2 \pm 23.6$} & \multicolumn{1}{c|}{$\mathbf{79.3 \pm 24.7}$} & $\mathbf{73.5 \pm 23.2}$ \\ \hline
\multicolumn{1}{|c|}{StrategicAI, $C=0$ (ours)} & \multicolumn{1}{c|}{$\mathbf{79.6 \pm 24.3}$} & \multicolumn{1}{c|}{$\mathbf{77.6 \pm 24.7}$} & \multicolumn{1}{c|}{$75.5 \pm 25.2$} & $66.6 \pm 23.3$ \\ \hline
\multicolumn{1}{|c|}{GreedyAI} & \multicolumn{1}{c|}{$ 71.0 \pm 22.6 $} & \multicolumn{1}{c|}{$ 69.4 \pm 22.5 $} & \multicolumn{1}{c|}{$ 69.5 \pm 22.6 $} & $ 67.5 \pm 21.6 $ \\ \hline
\multicolumn{1}{|c|}{RandomAI} & \multicolumn{1}{c|}{$ 71.0 \pm 20.8 $} & \multicolumn{1}{c|}{$ 64.2 \pm 20.3 $} & \multicolumn{1}{c|}{$ 62.0 \pm 21.2 $} & $ 58.7 \pm 20.0 $ \\ \hline
\multicolumn{1}{|c|}{Random} & \multicolumn{4}{c|}{$ 52.3 \pm 9.1 $} \\ \hline
\end{tabular}
\end{table}

\subsubsection{Experiment 3: Impact of the prior knowledge allocation.} 
Table~\ref{tab:impact-prior} shows the impact of the prior knowledge allocation on the optimization score when all the $(\alpha,\beta)$ configurations are combined. The results reveal that the AI's strategic planning improves the optimization performance regardless of the agent and the quality of prior knowledge they possess about the function. The only exception occurs when both agents lack prior knowledge. This may harm the initialization of the AI's own Gaussian process belief. In such cases, early-round planning becomes ineffective.
It is worth mentioning that the performance gap between the strategic AI agent and the greedy AI agent is usually most significant when the AI agent possesses high-quality prior information, as demonstrated by the results in rows 1-3 of Table~\ref{tab:impact-prior}.

\begin{table}[!b]
\caption{Impact of the agents' prior knowledge onto the optimization score. The tested priors are: knowledge around the global optimum (G), knowledge around a local optimum (L) and no prior knowledge (N). Each prior condition is indicated with a subscript: \textit{AI} for the AI agent, \textit{u} for the user.}
\label{tab:impact-prior}
\setlength{\tabcolsep}{10pt}
\def\arraystretch{1.2}
\centering
\begin{tabular}{|c|c|c|}
\hline
Prior & StrategicAI (ours) & GreedyAI \\ \hline
G$_{AI}$ \& G$_{u}$ & $ \mathbf{76.3 \pm 23.3} $ &      $ 63.6 \pm 18.6 $    \\ \hline
G$_{AI}$ \& L$_{u}$ & $ \mathbf{75.4 \pm 23.7} $ &     $ 67.4 \pm 23.2 $     \\ \hline
G$_{AI}$ \& N$_{u}$ & $ \mathbf{74.0 \pm 25.1} $ &       $ 61.3 \pm 18.6 $   \\ \hline
L$_{AI}$ \& G$_{u}$ & $ \mathbf{79.0 \pm 23.0} $ &     $ 75.8 \pm 22.7 $     \\ \hline
L$_{AI}$ \& L$_{u}$ & $ \mathbf{82.1 \pm 23.8} $ &       $ 70.1 \pm 25.5 $   \\ \hline
L$_{AI}$ \& N$_{u}$ & $ \mathbf{80.0 \pm 23.5} $ &      $ 75.1 \pm 23.6 $    \\ \hline
N$_{AI}$ \& N$_{u}$ & $ 69.5 \pm 23.3 $ &      $ \mathbf{72.1 \pm 20.4} $    \\ \hline
\end{tabular}
\end{table}

\subsubsection{Experiment 4: User certainty about the global maximum.} 

The optimization score alone may not provide a complete picture of the performance of collaboration, as the team may achieve a high function value but not ``know" whether it is indeed close to the global maximum. Such certainty requires knowledge of the overall domain, which in turn necessitates exploration. To assess the level of exploration and knowledge, we examine the flatness of the distribution of the maximum based on the agent's belief over the function, represented as $p(z^*|f_{\textrm{belief}}) := p(z^*=\max_{(x,y)}f_{\textrm{belief}}(x,y))$. The degree of flatness is measured by differential entropy. Specifically, we are interested in how effectively the AI agent can increase the user's certainty about the maximum, which we refer to as the \textit{user certainty}, $H(p(z^*|f_{u}))$, where $H$ is the differential entropy and $f_{u}$ is the human user's belief over the objective function. A higher user certainty value means that the human user has a better understanding of the global maximum.

Table~\ref{tab:user-max-entropy} replicates Experiment~\ref{exp_user_parameters}, but instead of presenting the optimization score, it shows the user certainty about the global maximum. The results reveal that the AI using a random strategy is the most effective approach to reducing the user's uncertainty about the global maximum, and that there is a considerable amount of unexplored space left after $T=20$ rounds when AI acts strategically or greedily. However, the results also indicate that with strategic planning, the user's understanding of the global maximum is slightly improved, regardless of whether they are conservative or Bayesian users and whether they are explorative or exploitative. In addition, the observation that strategic planning enables users to explore more space is also supported by a visual inspection of some of the experimental trials, which can be found in the appendix.


\begin{table}[]
\caption{User certainty about the global maximum at the end of the game.}
\label{tab:user-max-entropy}
\setlength{\tabcolsep}{10pt}
\def\arraystretch{1.2}
\centering
\begin{tabular}{c|cccc|}
\cline{2-5}
 & \multicolumn{2}{c|}{$\beta = 0.2$} & \multicolumn{2}{c|}{$\beta = 0.7$} \\ \cline{2-5} 
 & \multicolumn{1}{c|}{$\alpha = 0.1$} & \multicolumn{1}{c|}{$\alpha = 0.6$} & \multicolumn{1}{c|}{$\alpha = 0.1$} & $\alpha = 0.6$ \\ \hline
\multicolumn{1}{|c|}{StrategicAI} & \multicolumn{1}{c|}{$1.54 \pm 0.13$} & \multicolumn{1}{c|}{$1.59 \pm 0.15$} & \multicolumn{1}{c|}{$1.56 \pm 0.17$} & $1.56 \pm 0.19$ \\ \hline
\multicolumn{1}{|c|}{GreedyAI} & \multicolumn{1}{c|}{$ 1.66 \pm 0.05 $} & \multicolumn{1}{c|}{$ 1.67 \pm 0.06 $} & \multicolumn{1}{c|}{$ 1.68 \pm 0.06 $} & $ 1.68 \pm 0.06 $ \\ \hline
\multicolumn{1}{|c|}{RandomAI} & \multicolumn{1}{c|}{$ \mathbf{1.27 \pm 0.17} $} & \multicolumn{1}{c|}{$ \mathbf{1.22 \pm 0.19} $} & \multicolumn{1}{c|}{$ \mathbf{1.16 \pm 0.22 }$} & $ \mathbf{1.17 \pm 0.22} $ \\ \hline
\end{tabular}
\end{table}

\section{Related work}
\paragraph{Decomposition-based optimization.} 
The proposed cooperative BO game resembles a decomposition-based optimizer. Decomposed optimization partitions the dimensions of the optimized function into disjoint subsets and optimizes separately over these partitions \cite{duan2019cooperative}. Two popular families of decomposition-based optimizers are coordinate descent based methods \cite{hildreth1957} and cooperative co-evolutionary algorithms \cite{potter1994cooperative}. Recently, \cite{jiang2022cooperative} proposed a decomposition-based optimization algorithm for large-scale optimization problems, which is based on Bayesian optimization. However, this literature is focused on algorithmic optimization and does not address the problem from the multi-agent learning perspective.

\paragraph{Multi-agent Bayesian Optimization.} 
The closest to our work is collaborative BO which considers multiple parties optimizing the same objective function. Still, the utility from evaluating the function is individual, as \cite{sim2021collaborative}, where a trusted mediator selects an input query to be assigned to each party who then evaluates the objective function at the assigned input. The main difference with our work is the absence of this mediator. In other words, in our setting, the parties have autonomy over their own decisions. 

\paragraph{Human-Agent Teaming.}
The autonomy mentioned above is a crucial characteristic of human–autonomy teams (HATs), where autonomous agents with a partial or high degree of self-governance work toward a common goal \cite{larson2020leading,o2022human}. The HAT literature offers numerous testbeds that enable researchers to design algorithms and evaluate performance; a selection of these is presented in \cite{o2022human}. One such testbed is the game of Hanabi, which is a cooperative card game of imperfect information for two to five players \cite{bard2020hanabi}. Although the proposed cooperative BO game is similar to Hanabi, the crucial difference is that we do not allow direct communication, which would make collaboration easier and focus the solution on designing the communication aspects. By contrast, in Hanabi, players can exchange hints as a means of communication. This idea of communication is inherent to the whole field of Cooperative Game Theory~\cite{chalkiadakis2012cooperative}, in which cooperation is made possible by using binding agreements. However, this domain mainly focuses on matrix games and not sequential repeated games in extensive form. Recently, Sundin et al.~\cite{sundin2022human} considered a similar problem for an application to molecular design. In this work, the first agent's action corresponds to a restriction of a search space, and the second agent's action to picking within the restricted space. This differs from our work in that the function they optimize is known by the second agent but not observed by the first agent, while we consider a function unknown by both agents and the samples of which are observed.

\section{Conclusion}
We introduced a cooperative setup for Bayesian optimization of a function of two parameters, where a user and an AI agent sequentially select one coordinate each. The case where the AI agent chooses first is difficult because the agent cannot know the user's action. Therefore, we endow the AI agent with a model of the user, i.e. a probabilistic description of the user's behaviour and decision-making. We use this model within a Bayes Adaptive Monte Carlo Planning algorithm to simulate the user's behaviour. The AI agent's strategic planning of actions enables making choices adapted to the user's biases and current knowledge of the domain. 
We showed empirically that our method, based on a simple user model, leads to better optimization scores than a non-strategic planner. 
Even though our algorithm is, in principle, adapted to be used with human users, the current implementation is yet too computationally expensive to work in real-time (calculation time of the order of a minute per action). Alleviating this issue is an important future work to make our method usable in real-world applications with real users.

\section*{Acknowledgements}

This research was supported by EU Horizon 2020 (HumanE AI NET, 952026) and UKRI Turing AI World-Leading Researcher Fellowship (EP/W002973/1). Computational resources were provided by the Aalto Science-IT project from Computer Science IT. The authors would like to thank Prof. Frans Oliehoek and Dr. Mert Celikok for their help in setting up the project and the reviewers for their insightful comments.

%
%
%
\bibliographystyle{ECML-formatting/splncs04}
\bibliography{main}

\end{document}


\title{Supplementary Material for\\ Cooperative Bayesian optimization for imperfect agents
}

\author{Ali Khoshvishkaie\inst{1} \and Petrus Mikkola\inst{1} \and Pierre-Alexandre Murena\inst{1,2} \and Samuel Kaski\inst{1,3}}

\authorrunning{A. Khoshvishkaie et al.}
%

\institute{Department of Computer Science, Aalto University, Helsinki, Finland \and Hamburg University of Technology, Hamburg, Germany \and Department of Computer Science, University of Manchester, Manchester, UK \\ 
\email{\{firstname.lastname\}@aalto.fi}}

\maketitle              

\section{Experiment 1: Impact of a known user model}

Table~\ref{tab:impact-true-user-model} compares our strategic planning performance scores to an unrealistic case when having a perfect user model for all prior initialization conditions. Surprisingly, our method is performing comparably well, showing that the user model is able to capture essential characteristics of the user eventually.

\begin{table}[]
\caption{Impact of the knowing the user's prior knowledge and parameters onto the optimization score. The tested priors are: knowledge around the global optimum (G), knowledge around a local optimum (L) and no prior knowledge (N). Each prior condition is indicated with a subscript: \textit{AI} for the AI agent, \textit{u} for the user.}
\label{tab:impact-true-user-model}
\setlength{\tabcolsep}{10pt}
\def\arraystretch{1.2}
\centering
\begin{tabular}{|c|c|c|}
\hline
Prior & StrategicAI (ours) & StrategicAI+U \\ \hline
G$_{AI}$ \& G$_{u}$ & $ \mathbf{76.3 \pm 23.3} $ &      $ 72.0 \pm 22.2 $    \\ \hline
G$_{AI}$ \& L$_{u}$ & $ \mathbf{75.4 \pm 23.7} $ &     $ 73.9 \pm 25.0 $     \\ \hline
G$_{AI}$ \& N$_{u}$ & $ \mathbf{74.0 \pm 25.1} $ &       $ 70.8 \pm 24.7 $   \\ \hline
L$_{AI}$ \& G$_{u}$ & $ {79.0 \pm 23.0} $ &     $ \mathbf{81.7 \pm 22.4} $     \\ \hline
L$_{AI}$ \& L$_{u}$ & $ {82.1 \pm 23.8} $ &       $ \mathbf{83.7 \pm 23.9} $   \\ \hline
L$_{AI}$ \& N$_{u}$ & $ {80.0 \pm 23.5} $ &      $ \mathbf{80.7 \pm 23.9} $    \\ \hline
N$_{AI}$ \& N$_{u}$ & $ 69.5 \pm 23.3 $ &      $ \mathbf{71.2 \pm 24.7} $    \\ \hline
\end{tabular}
\end{table}

\section{Experiment 2: Full Variability Study}

Table~\ref{tab:impact-prior-full} presents the performance scores for all prior initialization conditions, and for all values of the parameters of the user model. 

\begin{table}[]
\caption{Impact of the agents' prior knowledge onto the optimization score. The tested priors are: knowledge around the global optimum (G), knowledge around a local optimum (L) and no prior knowledge (N). Each prior condition is indicated with a subscript: \textit{AI} for the AI agent, \textit{u} for the user.}
\label{tab:impact-prior-full}
\setlength{\tabcolsep}{3pt}
\def\arraystretch{1.2}
\centering
\begin{tabular}{cc|cc|cc|}
\cline{3-6}

                                              &                      & \multicolumn{2}{c|}{$ \beta=0.2 $}       & \multicolumn{2}{c|}{$ \beta=0.7 $}                                            \\ \hline
\multicolumn{1}{|c|}{Prior}                   & Method               & \multicolumn{1}{c|}{$ \alpha=0.1 $ }                     &            $ \alpha=0.6 $           & \multicolumn{1}{c|}{$ \alpha=0.1 $}                     &  $ \alpha=0.6 $                    \\ \hline
\multicolumn{1}{|c|}{\multirow{2}{*}{G$_{AI}$ \& G$_{u}$}} & \textbf{StrategicAI} & \multicolumn{1}{c|}{\textbf{77.2 ± 21.4}} & \textbf{72.6 ± 20.4} & \multicolumn{1}{c|}{\textbf{80.8 ± 19.1}} & \textbf{74.6 ± 18.6} \\ \cline{2-6} 
\multicolumn{1}{|c|}{}                        & GreedyAI             & \multicolumn{1}{c|}{63.0 ± 15.5}          & 64.2 ± 17.3          & \multicolumn{1}{c|}{63.3 ± 13.9}          & 64.0 ± 15.4          \\ \hline
\multicolumn{1}{|c|}{\multirow{2}{*}{G$_{AI}$ \& L$_{u}$}} & \textbf{StrategicAI} & \multicolumn{1}{c|}{\textbf{68.3 ± 21.5}} & \textbf{76.4 ± 21.4} & \multicolumn{1}{c|}{\textbf{79.3 ± 21.4}} & \textbf{77.8 ± 17.2} \\ \cline{2-6} 
\multicolumn{1}{|c|}{}                        & GreedyAI             & \multicolumn{1}{c|}{66.9 ± 21.3}          & 67.3 ± 20.5          & \multicolumn{1}{c|}{68.4 ± 20.1}          & 67.1 ± 20.4          \\ \hline
\multicolumn{1}{|c|}{\multirow{2}{*}{G$_{AI}$ \& N$_{u}$}} & \textbf{StrategicAI} & \multicolumn{1}{c|}{\textbf{80.1 ± 21.3}} & \textbf{69.7 ± 21.8} & \multicolumn{1}{c|}{\textbf{72.9 ± 24.3}} & \textbf{73.2 ± 21.4} \\ \cline{2-6} 
\multicolumn{1}{|c|}{}                        & GreedyAI             & \multicolumn{1}{c|}{63.0 ± 15.5}          & 57.9 ± 16.6          & \multicolumn{1}{c|}{64.0 ± 17.4}          & 60.3 ± 13.9          \\ \hline

\multicolumn{1}{|c|}{\multirow{2}{*}{L$_{AI}$ \& G$_{u}$}} & \textbf{StrategicAI} & \multicolumn{1}{c|}{\textbf{80.8 ± 19.9}} & \textbf{78.1 ± 18.5} & \multicolumn{1}{c|}{\textbf{81.8 ± 19.3}} & \textbf{75.3 ± 19.5} \\ \cline{2-6} 
\multicolumn{1}{|c|}{}                        & GreedyAI             & \multicolumn{1}{c|}{77.2 ± 19.3}          & 75.1 ± 20.4          & \multicolumn{1}{c|}{76.8 ± 19.1}          & 73.9 ± 18.3          \\ \hline
\multicolumn{1}{|c|}{\multirow{2}{*}{L$_{AI}$ \& L$_{u}$}} & \textbf{StrategicAI} & \multicolumn{1}{c|}{\textbf{83.5 ± 20.0}} & \textbf{84.0 ± 20.3} & \multicolumn{1}{c|}{\textbf{85.8 ± 18.8}} & \textbf{75.2 ± 20.6} \\ \cline{2-6} 
\multicolumn{1}{|c|}{}                        & GreedyAI             & \multicolumn{1}{c|}{71.2 ± 23.5}          & 69.3 ± 22.2          & \multicolumn{1}{c|}{70.9 ± 23.5}          & 69.2 ± 22.5          \\ \hline
\multicolumn{1}{|c|}{\multirow{2}{*}{L$_{AI}$ \& N$_{u}$}} & \textbf{StrategicAI} & \multicolumn{1}{c|}{\textbf{80.2 ± 20.8}} & \textbf{79.7 ± 17.4} & \multicolumn{1}{c|}{\textbf{80.4 ± 20.3}} & \textbf{79.8 ± 20.6} \\ \cline{2-6} 
\multicolumn{1}{|c|}{}                        & GreedyAI             & \multicolumn{1}{c|}{76.2 ± 20.7}          & 72.2 ± 19.0          & \multicolumn{1}{c|}{77.0 ± 21.3}          & 75.1 ± 20.3          \\ \hline
\multicolumn{1}{|c|}{\multirow{2}{*}{N$_{AI}$ \& N$_{u}$}} & \textbf{StrategicAI} & \multicolumn{1}{c|}{\textbf{72.6 ± 22.7}} & \textbf{73.2 ± 18.2} & \multicolumn{1}{c|}{\textbf{74.0 ± 21.7}} & \textbf{58.2 ± 18.9} \\ \cline{2-6} 
\multicolumn{1}{|c|}{}                        & GreedyAI             & \multicolumn{1}{c|}{79.6 ± 16.6}          & 79.6 ± 17.0          & \multicolumn{1}{c|}{66.2 ± 17.9}          & 63.1 ± 16.0          \\ \hline
\end{tabular}
\end{table}

\section{Acquisition trajectories}

In order to illustrate our method, we display here two acquisition trajectories, i.e. trajectories $\lbrace (x_t, y_t) \rbrace$ in the domain~$\mathcal{X} \times \mathcal{Y}$ of function $f$.

\begin{figure}[!h]
    \centering
    \includegraphics[width=.95\textwidth]{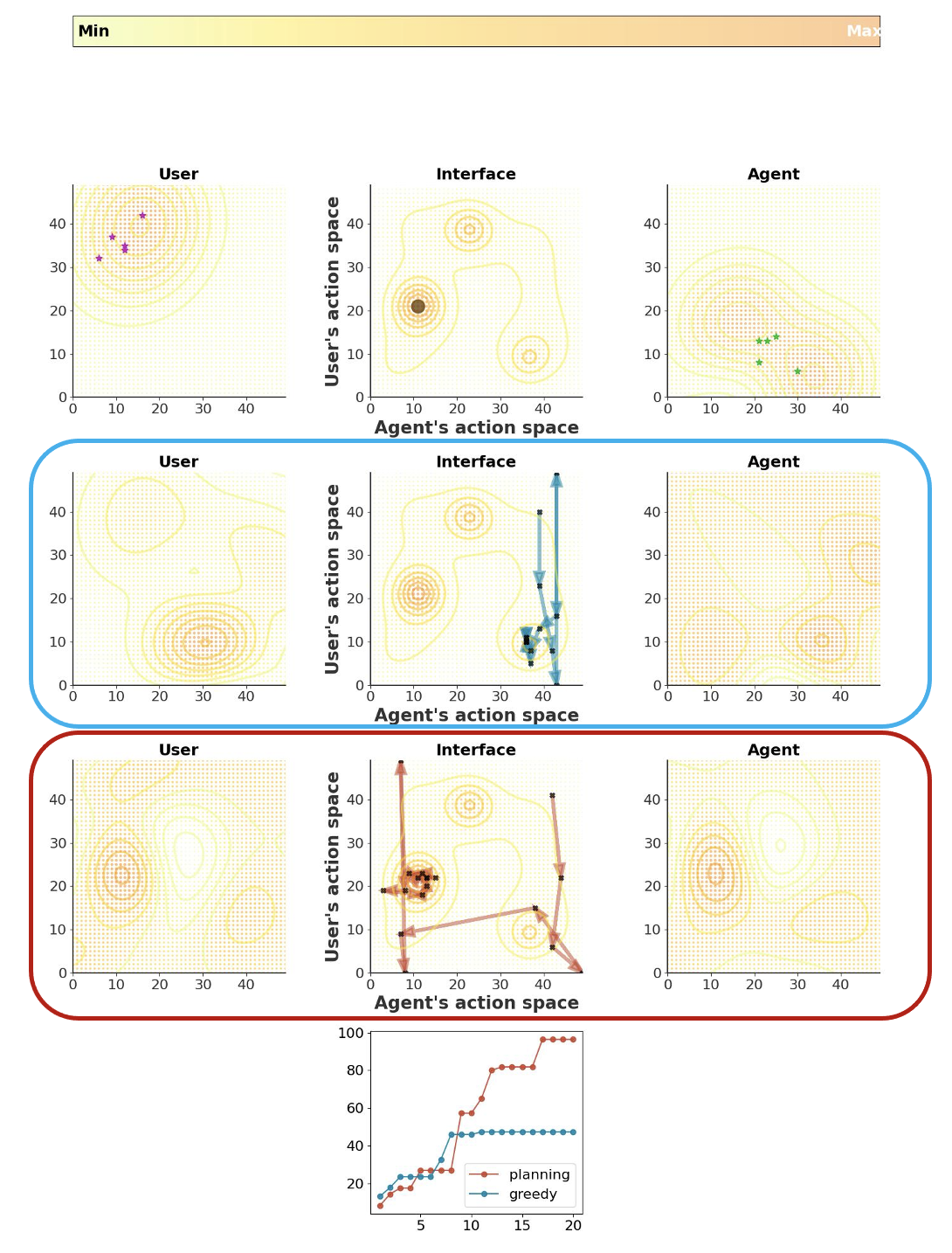}
    \caption{Trajectory of query points during an interaction. 
    The first column shows the user's knowledge of the function (row 1: initial knowledge, gained from prior observations marked in purple; row 2 and 3: final knowledge), and third column the AI agent's knowledge of the function. The second column displays real values of the function. The second and third rows demonstrate the scenarios of interaction between a specific user with GreedyAI and StrategicAI, respectively. The final subplot compares the optimization scores during the interaction.
    The strategic AI obviously outperforms the GreedyAI, helping the user explore the environment better and find the global optima. However, the the optimization performed with GreedyAI gets stuck at local optima and get a significantly lower score.}
    \label{fig:mainresult}
\end{figure}

\begin{figure}[!h]
    \centering
    \includegraphics[width=.95\textwidth]{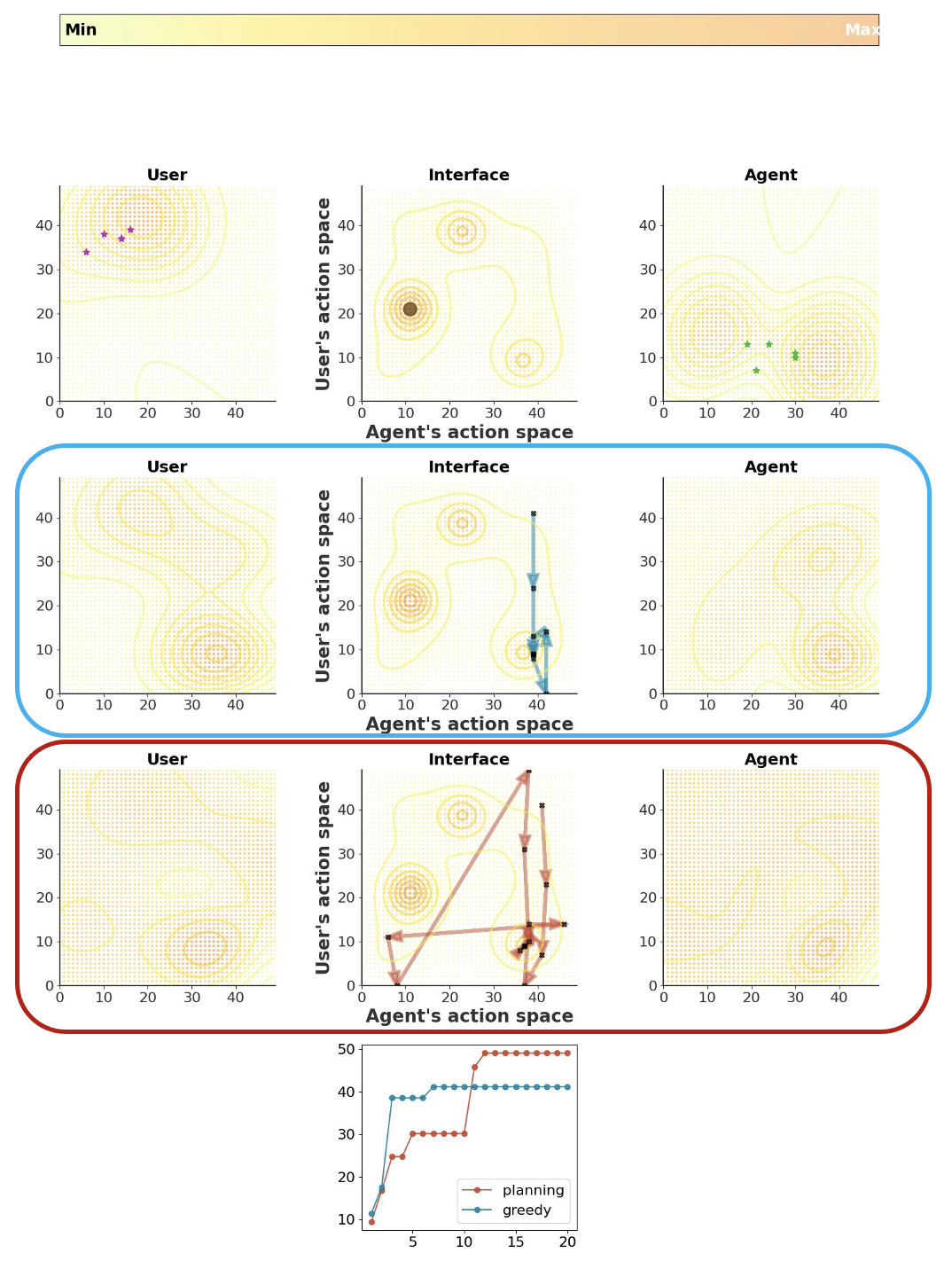}
    \caption{Trajectory of query points during an interaction. 
    The first column shows the user's knowledge of the function (row 1: initial knowledge, gained from prior observations marked in purple; row 2 and 3: final knowledge), and third column the AI agent's knowledge of the function. The second column displays real values of the function. The second and third rows demonstrate the scenarios of interaction between a specific user with GreedyAI and StrategicAI, respectively. The final subplot compares the optimization scores during the interaction.
    The StrategicAI explores the function better and outperforms the GreedyAI. However, both fail to find the global optimum.
    }
    \label{fig:mainresult}
\end{figure}

